\documentclass[10pt,twocolumn,letterpaper]{article}

\usepackage[pagenumbers]{cvpr} 

\definecolor{cvprblue}{rgb}{0.21,0.49,0.74}
\usepackage[pagebackref,breaklinks,colorlinks,allcolors=cvprblue]{hyperref}

\usepackage{listings}
\usepackage{xcolor}
\usepackage{amsmath}
\usepackage{wrapfig}
\usepackage{enumitem}
\usepackage{caption}
\usepackage{amsfonts}       
\usepackage{nicefrac}       
\usepackage{microtype}      
\usepackage{amsthm}
\usepackage{multirow}
\newtheorem{lemma}{Lemma}

\def\confName{CVPR}
\def\confYear{2027}

\title{A Mutual Information Perspective on Combining Sample-Wise and Feature-Wise Contrastive Learning}

\author{
{\normalsize\bfseries
Mahdi Beitollahi\textsuperscript{1} \quad
Mehdi Setayesh\textsuperscript{1} \quad
Mohammad Hossein Moslemi\textsuperscript{1} \quad
Hongliang Li\textsuperscript{1} \quad
Zhitang Chen\textsuperscript{1}
}
}

\begin{document}
\maketitle

\begingroup
\renewcommand{\thefootnote}{}
\footnotetext{
\textsuperscript{1}Noah’s Ark Lab
Correspondence to: Mahdi Beitollahi
\texttt{mahdi.beitollahi@queensu.ca}.
}
\endgroup

\begin{abstract}
Contrastive representation learning typically contrasts corresponding samples across two views, as in InfoNCE. However, the mutual information bound of sample-wise InfoNCE is limited by the number of negative samples and does not explicitly discourage redundancy across features. To address these issues, we study sample-wise and feature-wise contrastive learning through a common mutual information formulation to understand how to combine them. Our analysis of the feature axis reveals an additional penalty arising from the dependence and non-exchangeability of feature channels and motivates feature standardization before contrastive matching. The two objectives are complementary: the sample-wise objective distinguishes inputs, while the feature-wise objective promotes informative, non-redundant features. We prove that the mutual information captured along either axis lower-bounds the mutual information between the full embedding matrices, providing a principled basis for combining the two objectives. Our extensive experiments across image representation learning and cross-modal retrieval demonstrate that the combined objective improves over either objective alone and outperforms existing self-supervised learning methods in frozen-representation evaluations.
\end{abstract}

\section{Introduction} \label{sec}

Contrastive representation learning seeks to learn representations by making the representations of two different views of the same input similar, while making them distinguishable from representations of other inputs. A widely used objective is InfoNCE~\citep{oord2018representation}, where the representation of one view is treated as a query, and the model must identify the representation of its corresponding view among a set of representations from the same batch. Representations corresponding to other inputs act as negative examples, providing alternatives that the model is encouraged to distinguish from. Although effective, this construction imposes two limitations. First, because the set of alternatives is formed from other samples in the batch, the amount of mutual information that the objective can certify is limited by the number of negatives, creating a dependence on batch size~\citep{poole2019variational,mcallester2020formal,lee2025similarities}. Second, the objective operates entirely at the level of individual samples: it encourages representations of different inputs to be distinguishable, but places no direct constraint on how information is distributed across the feature channels within each representation. Consequently, different output channels may remain redundant, potentially wasting representational capacity and contributing to dimensional collapse~\citep{sansone2025collapse}.

The feature dimension provides another perspective on contrastive representation learning. Rather than distinguishing different inputs, we can distinguish the output feature channels by asking whether each channel can identify its counterpart across two
views. This introduces a feature-level matching signal that is not explicitly present in conventional sample-wise InfoNCE. In particular, the objective encourages corresponding feature channels to remain consistent across augmentations while making them distinguishable from other channels. Importantly, this does not require features to be mutually orthogonal; it imposes a relative matching criterion between corresponding and non-corresponding channels. Because the number of feature candidates is determined by the representation dimension rather than the minibatch size, this construction also provides
an alternative contrastive signal when the number of available sample negatives is limited.

Several approaches address the limitations of InfoNCE from different directions. BYOL~\citep{grill2020bootstrap} and SimSiam~\citep{chen2021exploring} avoid explicit negative samples through architectural asymmetry, reducing the dependence on large batches. Barlow Twins~\citep{zbontar2021barlow} and VICReg~\citep{bardes2021vicreg}, in contrast, explicitly regularize feature channels to encourage non-redundant representations. Feature-wise contrastive objectives have also been explored, including DimCL~\citep{nguyen2023dimcl} and SimCSE++~\citep{xu2023simcse++}, primarily to promote feature diversity or address feature redundancy. These approaches highlight two distinct aspects of representation learning: distinguishing different samples and distributing information across different features. However, the sample-wise and feature-wise contrastive perspectives have not been characterized together through the mutual information formulation underlying InfoNCE, and it remains unclear how they can be combined in a principled manner.

\begin{figure*}[t]
\centering
\includegraphics[width=\textwidth, trim={1cm 0cm 1cm 0.4cm}, clip]{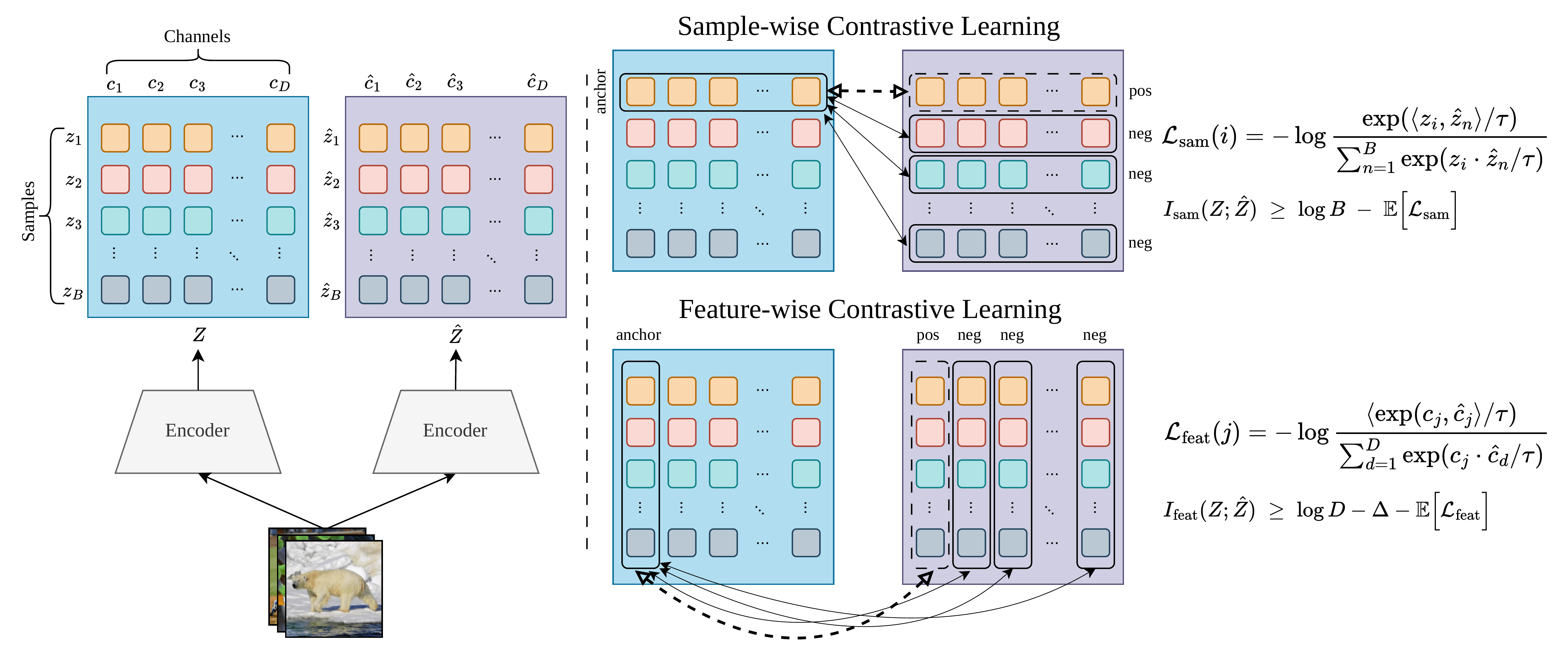}
\caption{Sample-wise contrastive learning contrasts rows over $B$ samples, while feature-wise contrastive learning contrasts columns over $D$ features, yielding mutual information bounds governed by $\log B$ and $\log D-\Delta$, respectively.}
\label{fig:sysmodel}
\end{figure*}

We therefore study the sample and feature axes through a common mutual information formulation to understand how to combine them, as shown in Figure~\ref{fig:sysmodel}. Let $Z,\hat Z\in\mathbb{R}^{B\times D}$ denote the representations produced by two branches, where $B$ indexes samples and $D$ indexes output features. The conventional sample-wise perspective treats the $B$ rows as the population and asks whether an embedding from one branch can identify its corresponding embedding among the samples of the other branch. The feature-wise perspective instead treats the $D$ columns as the population and asks whether a feature channel can identify its corresponding channel among those of the other representation.

Applying the mutual information formulation along the feature axis yields a feature-wise contrastive objective closely related to those used in prior work. Our analysis, however, reveals an additional penalty arising from the dependence and non-exchangeability of feature candidates, a distinction that does not appear in the standard sample-wise formulation. This analysis motivates feature standardization before contrastive matching. It also shows that the feature-wise bound is governed by the feature dimension $D$ rather than the batch size, providing a complementary contrastive signal. Our experiments further highlight the importance of these design choices.

We connect the two axis-wise objectives to the mutual information of the full embedding matrices. We prove that both the sample-wise and feature-wise mutual information quantities lower-bound the joint mutual information between $Z$ and $\hat Z$. This establishes that the two axis-wise formulations provide complementary lower bounds on the same underlying quantity. Consequently, we combine them through a convex combination, obtaining a combined objective with a corresponding lower bound on the joint mutual information.

Our formulation also reveals a connection between contrastive learning and redundancy-reduction methods such as Barlow Twins. Although these approaches are typically presented as distinct, both encourage corresponding representations to agree while discouraging redundancy across features. Our analysis shows that they can be viewed within a common matching framework: our feature-wise contrastive objective enforces matching through relative competition among features, whereas Barlow Twins directly penalizes deviations of feature correlations from their desired values.

In summary, our work makes three main contributions: 1) we characterize the mutual information interpretation of feature-wise contrastive learning, identifying the additional dependence and non-exchangeability penalty that distinguishes it from standard sample-wise InfoNCE; 2) we establish a common mutual information perspective on the sample and feature axes, relating their respective
bounds to the mutual information of the full embedding matrices and clarifying the conditions under which these bounds are informative; and 3) we empirically investigate the behavior and complementarity of the two objectives, demonstrating improvements in frozen representation quality and analyzing their sensitivity to batch size, feature dimension, and other training parameters.

\section{Problem Formulation}\label{sec:problem} Let $\mathcal{D} = \{(x_n,\hat x_n)\}_{n=1}^{M}$ denote a dataset of $M$ paired observations drawn from two branches. The two branches may correspond to two augmented views of the same input, as is standard in self-supervised representation learning \citep{chen2020simple}, or to two different modalities \citep{radford2021learning}. Each branch is processed by its respective encoder, $\phi$ and $\hat\phi$ (which may share weights, as is typical in single-network multiview architectures \citep{chen2020simple, bardes2021vicreg}), mapping inputs into a shared $D$-dimensional output space. For a minibatch of $B$ paired observations, stacking their representations row-wise yields two matrices
$$
Z, \hat Z \in \mathbb{R}^{B \times D}, \qquad Z_{i,d} = \phi(x_i)_d, \quad \hat Z_{i,d} = \hat\phi(\hat x_i)_d,
$$
with $i \in [B]$ indexing samples and $d \in [D]$ indexing output features. The matrices $Z$ and $\hat Z$ admit two complementary statistical perspectives:

\begin{itemize}
\item \textbf{Sample-wise:} The statistical population consists of the $B$ samples, with each sample represented by its $D$ output features. We write $z_i := Z_{i,:}$ and $\hat z_i := \hat Z_{i,:}$ (both in $\mathbb{R}^D$) to denote the two embeddings of sample $i$. Let $I\sim\mathrm{Unif}[B]$ be independent of $(Z,\hat Z)$.

\item \textbf{Feature-wise:} The statistical population consists of the $D$ output features, with each feature represented by its values across the $B$ samples. We write $c_d := Z_{:,d}$ and $\hat c_d := \hat Z_{:,d}$ (both in $\mathbb{R}^B$) to denote the two representations of feature $d$. Let $J\sim\mathrm{Unif}[D]$ be independent of $(Z,\hat Z)$.
\end{itemize}

The corresponding sample-wise and feature-wise mutual informations are defined as
\begin{equation}
I_{\mathrm{sam}}(Z;\hat Z)
:=
I(z_I;\hat z_I\mid I)
=
\frac{1}{B}\sum_{i=1}^{B} I(z_i;\hat z_i),
\end{equation}
\begin{equation}
I_{\mathrm{feat}}(Z;\hat Z)
:=
I(c_J;\hat c_J\mid J)
=
\frac{1}{D}\sum_{j=1}^{D} I(c_j;\hat c_j).
\end{equation}
The sample-wise perspective corresponds to the conventional contrastive-learning formulation, where each sample embedding is matched to its corresponding embedding among candidate samples from the other branch. The feature-wise perspective instead treats each output feature as the unit of correspondence, asking whether a feature from one branch can be identified among the $D$ output features of the other branch. In the following sections, we derive contrastive lower bounds for both $I_{\mathrm{sam}}(Z;\hat Z)$ and $I_{\mathrm{feat}}(Z;\hat Z)$ and relate them to the joint mutual information $I(Z;\hat Z)$ between the two embedding matrices.

\section{Mutual Information Objectives}
We now derive contrastive variational bounds for the sample-wise and feature-wise mutual information quantities introduced in Section~\ref{sec:problem}, and use their tractable loss terms as training objectives.

\subsection{Sample-wise Contrastive Objective}
First, we formulate the sample-wise contrastive framework and derive the InfoNCE loss \citep{oord2018representation}, which subsequently enables the formulation of the feature-wise contrastive loss. Fix an anchor sample $z_i$ and its corresponding partner $\hat z_i$. Let $U\sim\mathrm{Unif}[B]$ be independent of $(z_i,\hat z_i)$, and construct a candidate tuple $X=(X_1,\ldots,X_B)$ by placing the true partner $\hat z_i$ in slot $U$ and drawing the remaining $B-1$ candidates independently from the marginal distribution of $\hat z$. Since the candidate slots can be re-indexed without changing the softmax objective, we write the resulting bound in the conventional indexing where the positive candidate is $\hat z_i$:
\begin{equation}
\begin{aligned}
I(z_i;\hat z_i)
&\ge I(U;z_i,X) \\
&\ge \log B
+\mathbb{E}\left[
\log
\frac{\exp f^\star(z_i,\hat z_i)}
{\sum_{n=1}^{B} \exp f^\star(z_i,\hat z_n)}
\right].
\end{aligned}
\label{eq:barber-agakov}
\end{equation}
where $f^\star(z,\hat z) = \log \frac{p(z,\hat z)}{p(z)\,p(\hat z)}$ is the true pointwise mutual information critic. The first inequality is proved, via the chain rule for mutual information, in Appendix~\ref{app:proof-outer-ineq}; the second inequality is the Barber--Agakov bound~\citep{barber2004algorithm}, which more generally holds for any scoring function.

In practice (e.g., InfoNCE), the scoring function is commonly parameterized by the cosine similarity between the embeddings,
$\langle z_i, \hat z_n \rangle :=
\left\langle
\frac{z_i}{\|z_i\|_2},
\frac{\hat z_n}{\|\hat z_n\|_2}
\right\rangle$,
with a temperature parameter $\tau > 0$. Re-indexing the candidate tuple so that the positive partner occupies slot \(i\) gives the standard sample-wise contrastive loss:
\begin{equation}
\mathcal{L}_{\text{sam}}(i)
=
-\log
\frac{
\exp\big(\langle z_i, \hat z_i \rangle/\tau\big)
}{
\displaystyle
\sum_{n=1}^{B}
\exp\big(\langle z_i, \hat z_n \rangle/\tau\big)
},
\label{eq:sample-loss}
\end{equation}
\begin{equation}
I_{\text{sam}}(Z;\hat Z)
\;\ge\;
\log B
-
\mathbb{E}\big[\mathcal{L}_{\text{sam}}\big].
\label{eq:sample-bound}
\end{equation}

Since $\mathcal{L}_{\text{sam}} \ge 0$, the lower bound in Equation~\ref{eq:sample-bound} cannot exceed $\log B$. Consequently, the batch size imposes a hard ceiling on the mutual information that the sample-wise InfoNCE bound can certify at each step. As demonstrated in our experiments and prior literature \citep{ash2021investigating, chen2020simple}, the performance of $\mathcal{L}_{\text{sam}}$ can therefore be sensitive to the batch size. Raising this ceiling requires increasing the batch size, which demands substantial computational resources. 

\subsection{Feature-wise Contrastive Objective}

The sample-wise construction treats rows of $Z$ as the population. We now apply the same mutual information identification principle to the columns of $Z$, treating output features as the population. Fixing an anchor feature $c_j$, we consider the $D-1$ channels $\hat c_d$ ($d\neq j$) as negative candidates. Let $V\in[D]$ denote the index of the true partner among the $D$ candidates after randomly relabeling their positions, and let $Y=(Y_1,\ldots,Y_D)$ denote the resulting candidate tuple, such that $Y_V=\hat c_j$. The random relabeling ensures that $V$ is uniform over $[D]$.

The feature-wise construction differs from the sample-wise case because feature channels are jointly generated and are not assumed to be conditionally independent. Specifically, $c_j$ and the candidate channels are produced from the same batch of inputs and are coupled through the shared data and encoder architecture. Consequently, unlike the sample-wise negatives, the negative feature channels cannot generally be treated as conditionally independent of the anchor given its true partner. Moreover, feature channels are not naturally exchangeable; for example, nothing guarantees that channel 7 plays a statistically interchangeable role with channel 12.

As a result, when we formulate the mutual information bound for the feature-wise setting, an additional penalty term survives the derivation (proved in Appendix~\ref{app:proof-noniid-penalty}):
\begin{equation}
\begin{aligned}
I(c_j;\hat c_j)
&\ge I(V;c_j,Y)-\Delta_j \\
&\ge \log D-\Delta_j
+\mathbb{E}\left[
\log
\frac{\exp f(c_j,\hat c_j)}
{\sum_{d=1}^{D}\exp f(c_j,\hat c_d)}
\right].
\end{aligned}
\label{eq:barber-agakov-feature}
\end{equation}
where the final expression is written using the original feature indexing and the feature-wise dependence penalty is defined as
\begin{equation}
\Delta_j
:=
I\big(c_j;Y_{-V}\mid Y_V,V\big)
+
I(V;Y)
\ge0,
\label{eq:feature-penalty}
\end{equation}
where $Y_{-V}:=(Y_d)_{d\neq V}$ denotes the tuple of $D-1$ negative channels.

The first term in $\Delta_j$ captures the conditional dependence between the anchor and the negative channels, while the second captures the non-exchangeability of the candidate channels.

\textbf{When is $\Delta$ small?}
The bound in Equation~\ref{eq:barber-agakov-feature} is informative when $\Delta$ is small relative to $\log D$. Its two terms capture two sources of deviation from the ideal feature-wise identification setting: dependence between the anchor and non-corresponding channels, and non-exchangeability among the candidate features. These quantities can be characterized more explicitly under a Gaussian representation model. Suppose the rows $(z_i,\hat z_i)$ are i.i.d.\ jointly Gaussian, and let $\Sigma_{\hat z}:=\operatorname{Cov}(\hat z_i)$ denote the within-branch covariance of a row representation. For standardized features, $\operatorname{tr}(\Sigma_{\hat z})=D$, and we show in Appendix~\ref{app:gaussian-delta} that
\begin{equation}
\Delta_j
\le
-\frac{B}{2}\log(1-R_j^2)
-\frac{B}{2}\log\det\Sigma_{\hat z},
\label{eq:gaussian-delta-standardized}
\end{equation}
where
$R_j^2:=1-\operatorname{Var}(Z_{i,j}\mid\hat z_i)/\operatorname{Var}(Z_{i,j}\mid\hat Z_{i,j})$
measures the additional dependence between the scalar feature $Z_{i,j}$ and the non-corresponding features $\hat Z_{i,-j}$ after conditioning on its matched feature $\hat Z_{i,j}$. Thus, $\Delta_j$ is small when the non-corresponding features provide little additional information beyond the matched feature and the within-branch covariance is close to isotropic. In particular, $R_j^2=0$ and $\Sigma_{\hat z}=I_D$ imply $\Delta_j=0$. These sufficient conditions motivate feature standardization, while Appendix~\ref{app:delta-synthetic} further examines the behavior of $\Delta$ in a controlled setting.

\textbf{Target objective.}
Motivated by this bound, we optimize the tractable feature-wise identification term $I(V;c_j,Y)$ through its variational classifier. Before computing the feature-wise objective, we standardize each feature channel across the batch to have zero mean and unit variance. We then parameterize the scoring function using cosine similarity between the standardized feature channels,
$\langle c_j,\hat c_d\rangle:=\left\langle\frac{c_j}{\|c_j\|_2},\frac{\hat c_d}{\|\hat c_d\|_2}\right\rangle$,
with temperature $\tau>0$, which yields
\begin{equation}
\mathcal{L}_{\mathrm{feat}}(j)
=
-\log
\frac{
\exp\big(\langle c_j,\hat c_j\rangle/\tau\big)
}{
\displaystyle
\sum_{d=1}^{D}
\exp\big(\langle c_j,\hat c_d\rangle/\tau\big)
},
\label{eq:feature-loss}
\end{equation}
\begin{equation}
I_{\mathrm{feat}}(Z;\hat Z)
\;\ge\;
\log D
-
\Delta
-
\mathbb{E}\big[\mathcal{L}_{\mathrm{feat}}\big]
,
\label{eq:feature-bound}
\end{equation}
where $\Delta := \frac{1}{D}\sum_{j=1}^{D}\Delta_j$.
Unlike the sample-wise contrastive bound, whose candidate-set size is determined by the minibatch size $B$, the feature-wise bound is governed by the feature dimension $D$. Since $\mathcal{L}_{\mathrm{feat}}\ge0$, the variational certificate in Equation~\ref{eq:feature-bound} cannot exceed $\log D-\Delta$. Consequently, the information ceiling associated with the number of contrastive candidates is controlled by $D$ instead of $B$, providing a complementary source of contrastive signal that is not directly limited by the number of samples in the minibatch.

\subsection{Combining the Objectives}
\label{sec:combining}

The sample-wise and feature-wise objectives provide variational lower bounds on two axis-wise mutual information quantities, $I_{\mathrm{sam}}(Z;\hat Z)$ and $I_{\mathrm{feat}}(Z;\hat Z)$, respectively. Although these quantities capture different aspects of the information shared between the two representations, both are bounded by the mutual information of the full embedding matrices.

\begin{lemma}[Axis-wise mutual information lower-bounds joint mutual information]
\label{lem:axis-lower-bound}
\[
I(Z;\hat Z) \ge I_{\mathrm{sam}}(Z;\hat Z),
\qquad
I(Z;\hat Z) \ge I_{\mathrm{feat}}(Z;\hat Z).
\]
\end{lemma}

The proof is provided in Appendix~\ref{app:proof-coverage}. Combining Lemma~\ref{lem:axis-lower-bound} with the sample-wise and feature-wise bounds gives
\begin{equation}
I(Z;\hat Z)
\ge
\max\left\{
\log B-\mathbb{E}\big[\mathcal L_{\mathrm{sam}}\big],
\;
\log D-\Delta-\mathbb{E}\big[\mathcal L_{\mathrm{feat}}\big]
\right\}.
\label{eq:2-bounds-con}
\end{equation}
Thus, the two axes provide distinct lower-bound certificates of the same joint mutual information, governed by the number of samples $B$ and output features $D$, respectively.

Equation~\ref{eq:2-bounds-con} provides the tighter certificate by taking the maximum of the two bounds. However, it is not directly suitable as a training objective because the feature-wise bound contains the representation-dependent penalty $\Delta$, which is not available as a tractable quantity during optimization. We therefore optimize the tractable loss terms from both objectives through the convex combination
\begin{equation}
\mathcal{L}_{\mathrm{comb}}
=
\pi \mathcal{L}_{\mathrm{sam}}
+
(1-\pi)\mathcal{L}_{\mathrm{feat}},
\qquad
\pi\in[0,1].
\label{eq:combined-loss}
\end{equation}

For any $\pi\in[0,1]$, the joint mutual information also satisfies
\begin{align}
I(Z;\hat Z)
&\ge
\pi\!\left(
\log B-\mathbb{E}[\mathcal L_{\mathrm{sam}}]
\right)
\nonumber\\
&\quad+
(1-\pi)\!\left(
\log D-\Delta-\mathbb{E}[\mathcal L_{\mathrm{feat}}]
\right)
\nonumber\\
&=
\pi\log B
+
(1-\pi)(\log D-\Delta) -\mathbb{E}\big[\mathcal L_{\mathrm{comb}}\big].
\label{eq:combined-bound}
\end{align}
The practical role of $\mathcal{L}_{\mathrm{comb}}$ is to preserve both tractable training signals: $\mathcal{L}_{\mathrm{sam}}$ distinguishes corresponding samples from other samples, while $\mathcal{L}_{\mathrm{feat}}$ distinguishes corresponding features from other features. The parameter $\pi$ controls the relative contribution of the two objectives, with $\pi=1$ and $\pi=0$ recovering the purely sample-wise and feature-wise objectives, respectively. Our experiments in Section~\ref{sec:exp-ablation} show that intermediate values of $\pi$ outperform either endpoint, supporting the empirical benefit of jointly optimizing the two axes. We analyze the computational and memory costs of the individual and combined objectives in Appendix~\ref{app:complexity}.

\begin{table*}[tbp]
\centering
\caption{ImageNet benchmark. $k$-NN, linear-probe, and full-fine-tuning Top-1/Top-5 accuracy (\%), together with training time and peak memory. Bold and underline indicate the best and second-best results, respectively.}
\label{tab:imagenet-results}
\renewcommand{\arraystretch}{1.2}
\resizebox{\textwidth}{!}{%
\begin{tabular}{l c c c c c c c c c}
\toprule
\multirow{2}{*}{Method}
& \multicolumn{2}{c}{$k$-NN}
& \multicolumn{2}{c}{Linear Probe}
& \multicolumn{2}{c}{Full Fine-Tuning}
& \multicolumn{3}{c}{Training Cost} \\
\cmidrule(lr){2-3}
\cmidrule(lr){4-5}
\cmidrule(lr){6-7}
\cmidrule(lr){8-10}
& Top-1 & Top-5
& Top-1 & Top-5
& Top-1 & Top-5
& Time / epoch & Peak memory & Relative time \\
\midrule
SimCLR
& 44.8 & 73.9
& 63.2 & 85.2
& 73.9 & \textbf{91.9}
& 29.2 min & 11.3 GB & 1.00$\times$ \\

MoCo
& 41.8 & 72.2
& 61.5 & 84.1
& \underline{74.3} & \textbf{91.9}
& 24.9 min & 6.8 GB & 0.85$\times$ \\

BYOL
& 46.0 & 74.8
& 62.5 & 85.0
& \textbf{74.5} & \textbf{91.9}
& 41.2 min & 11.7 GB & 1.41$\times$ \\

TiCo
& 26.6 & 53.6
& 49.7 & 74.4
& 72.7 & 90.9
& 22.4 min & 12.8 GB & 0.77$\times$ \\

Barlow Twins
& 45.6 & 73.9
& 62.9 & 84.3
& 72.6 & 90.9
& 37.9 min & 14.0 GB & 1.30$\times$ \\

VICReg
& 46.3 & 75.2
& 63.0 & 85.4
& 73.7 & \textbf{91.9}
& 40.1 min & 15.7 GB & 1.37$\times$ \\

MACL & 46.0 & 74.4 & 62.1 & 83.9 & 71.8 & 90.4 & 30.3 min & 11.3 GB & 1.03$\times$ \\

SimCLR + DimCL
& 45.1 & 74.1
& 63.4 & 85.4
& 73.1 & 90.9
& 30.3 min & 14.9 GB & 1.04$\times$ \\
\midrule

$\mathcal{L}_{\text{sam}}$
& 45.5 & 73.5
& 64.2 & 85.8
& 72.7 & 91.0
& 29.2 min & 13.6 GB & 1.00$\times$ \\

$\mathcal{L}_{\text{feat}}$
& \underline{50.1} & \underline{77.7}
& \underline{65.8} & \underline{86.8}
& 72.6 & 91.0
& 30.3 min & 14.9 GB & 1.04$\times$ \\

$\mathcal{L}_{\text{comb}}$
& \textbf{50.7} & \textbf{78.0}
& \textbf{67.2} & \textbf{87.6}
& 73.2 & \underline{91.3}
& 31.1 min & 14.9 GB & 1.07$\times$ \\

\bottomrule
\end{tabular}%
}
\vspace{-0.3 cm}
\end{table*}

\section{Relation to Barlow Twins}
\label{sec:barlow}

Prior work has shown that sample-contrastive and covariance-based dimension-contrastive objectives are closely related~\citep{garrido2023duality}. Our formulation provides a complementary mutual-information view by applying the InfoNCE identification principle directly along the feature axis. 
Specifically, Barlow Twins~\citep{zbontar2021barlow} encourages agreement between corresponding features while reducing redundancy between distinct features. Given standardized feature channels, it constructs the cross-correlation matrix $C_{dd'} := \langle c_d,\hat c_{d'}\rangle$, $C\in\mathbb{R}^{D\times D}$, where $C_{dd'}$ measures the cross-correlation between feature $d$ in one branch and feature $d'$ in the other. Our feature-wise contrastive objective operates on this same cross-branch feature-similarity matrix. In particular, Equation~\ref{eq:feature-loss} can be written as
\begin{equation}
\mathcal{L}_{\mathrm{feat}}(d)
=
-\frac{C_{dd}}{\tau}
+
\log\sum_{d'=1}^{D}
\exp\left(\frac{C_{dd'}}{\tau}\right).
\label{eq:feature-loss-C}
\end{equation}
This imposes a \emph{relative} matching criterion: the positive entry $C_{dd}$ is encouraged to dominate the competing entries $C_{dd'}$, $d'\neq d$, without prescribing their absolute values.

Barlow Twins instead applies an \emph{absolute} matching criterion to the same matrix:
\begin{equation}
\mathcal{L}_{\mathrm{BT}}
=
\sum_{d=1}^{D}(1-C_{dd})^2
+
\lambda
\sum_{d\neq d'} C_{dd'}^2,
\label{eq:feature-bt}
\end{equation}
which directly encourages $C$ to approach the identity matrix by driving corresponding-feature correlations toward one and cross-feature correlations toward zero.

Thus, the two objectives use the same cross-feature matching structure but enforce it differently: our feature-wise contrastive loss uses softmax competition, requiring each feature to be more similar to its counterpart than to alternatives, whereas Barlow Twins directly penalizes deviations from the target correlation structure. They therefore represent relative and absolute forms of cross-feature matching, respectively. More generally, absolute matching is not inherently tied to the feature axis; a Barlow Twins-style objective can analogously be defined over samples. Exploring absolute matching along the sample axis, as well as more general combinations of relative and absolute matching across both axes, is an interesting direction for future work.

\section{Experiments} \label{sec:experiments} We empirically investigate three questions motivated by our analysis: (i) whether feature-wise contrastive learning provides a useful representation-learning signal on its own; (ii) whether combining the sample-wise and feature-wise objectives improves over optimizing either axis alone; and (iii) whether the distinct dependence of the two objectives on the batch size $B$ and feature dimension $D$ is reflected in their empirical behavior.
\subsection{Experimental Setup} \label{sec:exp-setup} \textbf{Pretraining.} We follow the SimCLR framework~\citep{chen2020simple} and perform self-supervised pretraining on the ImageNet-1K (ILSVRC2012) training set~\citep{deng2009imagenet}. Unless otherwise stated, all methods use a ResNet-50 backbone~\citep{he2016deep} trained from scratch under the same data-augmentation and optimization pipeline. Our default projector output dimension is $D=8192$, the pretraining batch size is $B=256$, and the temperature is $\tau=0.1$. For the combined objective, we use $\pi=0.5$ unless otherwise specified. Complete architectural and optimization details are provided in Appendix~\ref{app:implementation}.

\textbf{Evaluation protocols.} We evaluate the learned representations using three complementary protocols. First, we report $k$-nearest-neighbor ($k$-NN) classification using representations extracted from the frozen backbone. Second, we perform linear evaluation by training a linear classifier on top of the frozen backbone while keeping all backbone parameters fixed. Third, we perform end-to-end fine-tuning of both the backbone and classification head. For ImageNet, we report Top-1 and Top-5 accuracy under all three protocols. The frozen $k$-NN and linear evaluations measure the quality of the representation produced directly by self-supervised pretraining, whereas full fine-tuning additionally measures how effectively the pretrained representation can be adapted using labeled data. 

\textbf{Baselines.} We compare against representative sample-wise contrastive, negative-free, and redundancy-reduction methods, including SimCLR~\citep{chen2020simple}, MoCo~\citep{he2020momentum}, BYOL~\citep{grill2020bootstrap}, TiCo~\citep{zhu2022tico},  MACL~\citep{huang2023model},  DimCL~\citep{nguyen2023dimcl}, Barlow Twins~\citep{zbontar2021barlow}, and VICReg~\citep{bardes2021vicreg}. We additionally compare the three variants of our formulation: the sample-wise objective $\mathcal{L}_{\mathrm{sam}}$, the feature-wise objective $\mathcal{L}_{\mathrm{feat}}$, and their combination $\mathcal{L}_{\mathrm{comb}}$.

\vspace{-0.1cm}
\subsection{ImageNet Representation Quality}
\label{sec:exp-imagenet}

We first evaluate whether feature-wise contrastive learning provides a useful representation-learning signal on its own and whether combining the sample-wise and feature-wise objectives is beneficial. As shown in Table~\ref{tab:imagenet-results}, $\mathcal{L}_{\mathrm{feat}}$ performs strongly under both $k$-NN and linear evaluation, demonstrating that contrasting feature channels can learn effective representations independently of a conventional sample-wise objective. Combining the two axes further improves frozen-representation performance and achieves the strongest results under both k-NN and linear evaluation, supporting the complementarity of the sample-wise and feature-wise signals. The differences are less pronounced after full fine-tuning, where supervised optimization can substantially reshape the pretrained representation and reduce the influence of the pretraining objective. The additional feature-wise computation introduces only a modest increase in training time and memory usage, with a detailed complexity analysis provided in Appendix~\ref{app:complexity}.

\subsection{Cross-Modal Retrieval} \label{sec:exp-retrieval} We evaluate whether the sample-wise/feature-wise formulation extends beyond two augmented views of the same modality. On Flickr30K~\citep{young2014image}, we instantiate the two branches using a frozen ViT-L/14 image encoder~\citep{dosovitskiy2020image} and a frozen BERT-base text encoder~\citep{devlin2019bert}, training only linear projection heads to isolate the effect of the alignment objective from representation learning within the modality-specific encoders. We evaluate both image-to-text and text-to-image retrieval using standard recall metrics. As shown in Table~\ref{tab:retrieval-flickr}, the feature-wise objective alone is less effective than in the single-modality setting, but combining the sample-wise and feature-wise objectives provides the strongest retrieval performance, suggesting that the two axes remain complementary even when the paired representations are produced by heterogeneous modalities and encoder architectures.

\begin{table}[htbp]
\centering
\caption{Cross-modal retrieval on Flickr30K~\citep{young2014image}. Bold marks the best entry in each column.}
\label{tab:retrieval-flickr}
\resizebox{0.95\columnwidth}{!}{%
\begin{tabular}{l ccc ccc}
\toprule
\multirow{2}{*}{Method} & \multicolumn{3}{c}{Image $\to$ Text} & \multicolumn{3}{c}{Text $\to$ Image} \\
\cmidrule(lr){2-4} \cmidrule(lr){5-7}
 & R@1 & R@5 & R@10 & R@1 & R@5 & R@10 \\
\midrule
VICReg & 9.63 & 25.29 & 36.52 & 9.00 & 24.82 & 34.79 \\
\midrule
$\mathcal{L}_{\text{sam}}$  & 8.02 & 23.47 & 33.66 & 7.82 & 23.53 & 34.10 \\
$\mathcal{L}_{\text{feat}}$ & 7.77 & 22.71 & 31.93 & 6.07 & 19.94 & 29.16 \\
$\mathcal L_{\text{comb}}$    & \textbf{9.75} & \textbf{28.09} & \textbf{39.32} & \textbf{10.47} & \textbf{27.12} & \textbf{38.85} \\
\bottomrule
\end{tabular}}
\end{table}

\begin{table*}[!htbp]
\centering
\footnotesize
\caption{$k$-NN, linear-probe, and full-fine-tuning Top-1 accuracy (\%) vs.\ pretraining batch size $B$.}
\label{tab:ablation-batchsize}
\renewcommand{\arraystretch}{1.2}
\resizebox{0.7\textwidth}{!}{%
\begin{tabular}{l ccc ccc ccc}
\toprule
\multirow{2}{*}{$B$} & \multicolumn{3}{c}{$k$-NN} & \multicolumn{3}{c}{Linear Probe} & \multicolumn{3}{c}{Full Fine-Tuning} \\
\cmidrule(lr){2-4} \cmidrule(lr){5-7} \cmidrule(lr){8-10}
 & $\mathcal{L}_{\text{sam}}$ & $\mathcal{L}_{\text{feat}}$ & $\mathcal L_{\text{comb}}$
 & $\mathcal{L}_{\text{sam}}$ & $\mathcal{L}_{\text{feat}}$ & $\mathcal L_{\text{comb}}$
 & $\mathcal{L}_{\text{sam}}$ & $\mathcal{L}_{\text{feat}}$ & $\mathcal L_{\text{comb}}$ \\
\midrule
64  & 37.2 & 46.4 & 46.8 & 60.4 & 63.8 & 64.1 & 69.9 & 71.8 & 72.1 \\
128 & 41.3 & 48.2 & 48.6 & 62.3 & 64.9 & 65.0 & 71.3 & 72.3 & 72.8 \\
256 & 45.5 & 50.1 & 50.7 & 64.2 & 65.8 & 67.2 & 72.7 & 72.6 & 73.2 \\
\bottomrule
\end{tabular}%
}
\end{table*}

\subsection{Analysis and Ablations}
\label{sec:exp-ablation}
We examine the effects of batch size $B$, feature dimension $D$,
combination weight $\pi$, and temperature $\tau$, and further
analyze feature dependence and mutual information.

\textbf{Effect of feature dimension.}
We vary the projector dimension $D$ while keeping other settings
fixed. As shown in Table~\ref{tab:featdim},
$\mathcal{L}_{\mathrm{sam}}$ is largely insensitive to $D$,
whereas $\mathcal{L}_{\mathrm{feat}}$ and
$\mathcal{L}_{\mathrm{comb}}$ benefit from larger dimensions,
consistent with the dependence of the feature-wise bound on $D$.

\begin{table}[htbp]
\centering
\footnotesize
\caption{$k$-NN and full fine-tuning Top-1 accuracy (\%) vs. projector output dimension $D$.}
\label{tab:featdim}
\renewcommand{\arraystretch}{1.2}
\resizebox{0.45\textwidth}{!}{%
\begin{tabular}{l ccc ccc}
\toprule
\multirow{2}{*}{$D$}
& \multicolumn{3}{c}{$k$-NN}
& \multicolumn{3}{c}{Full Fine-Tuning} \\
\cmidrule(lr){2-4}
\cmidrule(lr){5-7}
& $\mathcal{L}_{\mathrm{sam}}$
& $\mathcal{L}_{\mathrm{feat}}$
& $\mathcal{L}_{\mathrm{comb}}$
& $\mathcal{L}_{\mathrm{sam}}$
& $\mathcal{L}_{\mathrm{feat}}$
& $\mathcal{L}_{\mathrm{comb}}$ \\
\midrule
512
& 45.4 & 49.2 & 50.2
& 72.5 & 74.2 & 73.2 \\

2048
& 45.5 & 50.1 & 50.4
& 72.6 & 73.1 & 73.3 \\

8192
& 45.5 & 50.1 & 50.7
& 72.7 & 72.6 & 73.2 \\
\bottomrule
\end{tabular}%
}
\vspace{-0.1cm}
\end{table}

\begin{table}
\vspace{-0.1\baselineskip}
\centering
\small
\caption{ImageNet Top-1 accuracy (\%) as a function of the combination weight $\pi$ in $\mathcal{L}_{\mathrm{comb}}$.}
\label{tab:ablation-pi}
\resizebox{0.75\columnwidth}{!}{%
\begin{tabular}{@{}lccccc@{}}
\toprule
$\pi$ & 0.0 & 0.25 & 0.5 & 0.75 & 1.0 \\
\midrule
$k$-NN          & 50.1 & 50.7 & 50.7 & 48.6 & 45.5 \\
Linear          & 65.8 & 66.9 & 67.2 & 66.1 & 64.2 \\
Full fine-tune  & 72.6 & 73.3 & 73.2 & 72.9 & 72.7 \\
\bottomrule
\end{tabular}}
\vspace{-0.5\baselineskip}
\end{table}

\textbf{Complementarity of the two objectives.}
The combined objective in Equation~\ref{eq:combined-loss} interpolates between purely feature-wise and purely sample-wise learning through the weight $\pi$. We vary this weight while keeping all other training settings fixed to determine whether the two objectives provide complementary signals. Table~\ref{tab:ablation-pi} shows that intermediate mixtures outperform either single-axis objective, indicating that the benefit of the combined formulation arises from jointly exploiting the sample and feature axes rather than simply selecting the stronger individual objective.

\textbf{Sensitivity to batch size.} The sample-wise bound in Equation~\ref{eq:sample-bound} is directly governed by the number of samples in the minibatch, whereas the candidate-set term of the feature-wise bound is controlled by the feature dimension. We therefore vary the pretraining batch size while fixing $D$ and the remaining training configuration. As shown in Table~\ref{tab:ablation-batchsize}, the sample-wise objective improves substantially as the batch size increases, while the feature-wise objective is less sensitive to $B$ and performs comparatively better at smaller $B$. 

\textbf{Temperature sensitivity.}
As shown in Table~\ref{tab:ablation-temp}, all three
objectives are sensitive to $\tau$, with the strongest
frozen-representation performance at $\tau=0.1$.
The feature-wise objective is particularly sensitive to
small temperatures, while the combined objective
consistently outperforms either individual objective
under frozen evaluation.

\begin{table*}[!htbp]
\centering
\footnotesize
\caption{$k$-NN, linear-probe, and full-fine-tuning Top-1 accuracy (\%) vs.\ temperature $\tau$.}
\label{tab:ablation-temp}
\renewcommand{\arraystretch}{1.2}
\resizebox{0.7\textwidth}{!}{%
\begin{tabular}{l ccc ccc ccc}
\toprule
\multirow{2}{*}{$\tau$} & \multicolumn{3}{c}{$k$-NN} & \multicolumn{3}{c}{Linear Probe} & \multicolumn{3}{c}{Full Fine-Tuning} \\
\cmidrule(lr){2-4} \cmidrule(lr){5-7} \cmidrule(lr){8-10}
 & $\mathcal{L}_{\text{sam}}$ & $\mathcal{L}_{\text{feat}}$ & $\mathcal L_{\text{comb}}$
 & $\mathcal{L}_{\text{sam}}$ & $\mathcal{L}_{\text{feat}}$ & $\mathcal L_{\text{comb}}$
 & $\mathcal{L}_{\text{sam}}$ & $\mathcal{L}_{\text{feat}}$ & $\mathcal L_{\text{comb}}$ \\
\midrule
0.01 & 30.1 & 9.2 & 33.7 & 48.3 & 16.1 & 51.9 & 65.2 & 65.6 & 66.8 \\
0.1 & 45.5 & 50.1 & 50.7 & 64.2 & 65.8 & 67.2 & 72.7 & 72.6 & 73.2 \\
1.0  & 38.7 & 42.9 & 45.0 & 56.2 & 58.3 & 59.4 & 72.9 & 73.1 & 74.5 \\
\bottomrule
\end{tabular}%
}
\vspace{-0.3 cm}
\end{table*}

\textbf{Feature dependence and effective dimensionality.}
We next examine how the three objectives affect feature redundancy
and effective dimensionality in the learned backbone representations.
We evaluate the pretrained models on 10,000 held-out ImageNet-V2
images and measure the average absolute correlation between
different feature channels, the Gaussian total correlation
$-\log\det R$ normalized by the feature dimension, and
the effective rank using RankMe~\citep{garrido23a},
where $R$ denotes the feature correlation matrix.
As shown in Table~\ref{tab:feature-dependence},
both $\mathcal{L}_{\mathrm{feat}}$ and
$\mathcal{L}_{\mathrm{comb}}$ reduce feature dependence
compared with $\mathcal{L}_{\mathrm{sam}}$ while achieving
higher effective rank. In particular,
$\mathcal{L}_{\mathrm{feat}}$ reduces the average
channel correlation by approximately 23\%, while
$\mathcal{L}_{\mathrm{comb}}$ achieves the highest
effective rank. These results suggest that introducing
feature-wise contrastive learning encourages less
redundant and more diverse backbone representations.

\textbf{Contrastive matching along both axes.}
We further investigate whether combining the two objectives
improves the contrastive signals along both axes. For each
trained model, we compute the empirical scores
$S_{\mathrm{sam}}=\log B-\widehat{\mathcal{L}}_{\mathrm{sam}}$
and
$S_{\mathrm{feat}}=\log D-\widehat{\mathcal{L}}_{\mathrm{feat}}$
using held-out batches of size $B=256$ for the backbone representations.
As shown in Table~\ref{tab:feature-dependence},
$\mathcal{L}_{\mathrm{feat}}$ improves the feature-wise
score but reduces the sample-wise score compared with
$\mathcal{L}_{\mathrm{sam}}$. In contrast,
$\mathcal{L}_{\mathrm{comb}}$ achieves the highest
feature-wise score while maintaining the sample-wise
score of $\mathcal{L}_{\mathrm{sam}}$. This provides
further empirical evidence that the two objectives
capture complementary matching signals. Importantly,
the feature-wise score excludes the penalty $\Delta$
and therefore does not by itself constitute a certified
mutual-information lower bound.

\begin{table}[htbp]
\centering
\footnotesize
\caption{Feature dependence and effective rank of backbone
representations.}
\label{tab:feature-dependence}
\renewcommand{\arraystretch}{1.2}
\resizebox{0.95\columnwidth}{!}{%
\begin{tabular}{l ccc}
\toprule
Metric
& $\mathcal{L}_{\mathrm{sam}}$
& $\mathcal{L}_{\mathrm{feat}}$
& $\mathcal{L}_{\mathrm{comb}}$ \\
\midrule
\multicolumn{4}{l}{\textit{Backbone representations}} \\
Mean absolute correlation $\downarrow$
& 0.0580 & \textbf{0.0449} & 0.0482 \\
$-\log\det R / 2048$ $\downarrow$
& 1.16 & \textbf{0.99} & 1.02 \\
Effective rank (RankMe) $\uparrow$
& 1489 & 1531 & \textbf{1533} \\
\midrule
\multicolumn{4}{l}{\textit{Contrastive scores (nats)}} \\
$S_{\mathrm{sam}}$ $\uparrow$
& 5.21 & 4.52 & \textbf{5.22} \\
$S_{\mathrm{feat}}$ $\uparrow$
& 7.25 & 7.69 & \textbf{7.88} \\
\bottomrule
\end{tabular}%
}
\vspace{-0.2cm}
\end{table}

\textbf{Empirical evaluation of the feature-wise bound.}
We further investigate whether improving the feature-wise
contrastive objective corresponds to higher mutual information
between matching feature channels, and examine the dependence
penalty $\Delta$ in Equation~\ref{eq:feature-penalty}.
We evaluate the trained models on 10,000 held-out ImageNet-V2
images and estimate $I_{\mathrm{feat}}$ using both Gaussian
and nonparametric mutual information estimators.
As shown in Table~\ref{tab:delta-empirical},
the estimated feature-wise mutual information increases
from $\mathcal{L}_{\mathrm{sam}}$ to
$\mathcal{L}_{\mathrm{feat}}$ and
$\mathcal{L}_{\mathrm{comb}}$, consistent with the ordering
of their empirical feature-wise contrastive scores.
We additionally examine the two components of $\Delta$
using cross-fitted regression and anchor-free feature
identification probes. The regression probes detect no
additional dependence on non-matching channels beyond
the corresponding channel, while the identification
probes reveal limited detectable non-exchangeability.
Although these estimates do not provide an upper bound
on $\Delta$, the results show that improvements in the
feature-wise contrastive objective are accompanied by
higher estimated mutual information in the learned
representations.

\begin{table}[htbp]
\centering
\footnotesize
\caption{Estimated feature-wise mutual information and
diagnostic probes of $\Delta$. All values are
in nats.}
\label{tab:delta-empirical}
\renewcommand{\arraystretch}{1.2}
\resizebox{0.95\columnwidth}{!}{%
\begin{tabular}{l ccc}
\toprule
Metric
& $\mathcal{L}_{\mathrm{sam}}$
& $\mathcal{L}_{\mathrm{feat}}$
& $\mathcal{L}_{\mathrm{comb}}$ \\
\midrule
$I_{\mathrm{feat}}$ (KSG estimate)
& 131.7 & 207.6 & \textbf{224.0} \\
$I_{\mathrm{feat}}$ (Gaussian estimate)
& 125.8 & 178.5 & \textbf{199.5} \\
\midrule
$\Delta$: dependence (adjusted estimate)
& 2.8 & 7.2 & 2.4 \\
$\Delta$: non-exchangeability (probe lower bound)
& 0.02 & 0.30 & 0.22 \\
\bottomrule
\end{tabular}%
}
\vspace{-0.2cm}
\end{table}

\section{Related Work}
\textbf{Contrastive Learning and Mutual Information.}
Contrastive learning typically operates along the sample axis, with InfoNCE~\citep{oord2018representation} underlying methods such as MoCo~\citep{he2020momentum} and SimCLR~\citep{chen2020simple}. Its mutual-information interpretation and dependence on negative candidates have been extensively studied~\citep{poole2019variational,mcallester2020formal}. Recent work has further analyzed InfoNCE through its minibatch geometry~\citep{lee2025similarities}, identifiability~\citep{rusak2025infonce}, projection heads~\citep{ouyang2025projection}, probabilistic formulations~\citep{wang2025discriminative}, and relation to mutual-information estimation~\citep{ryu2026contrastive}, while related work has derived non-contrastive objectives from mutual-information principles~\citep{guo2026representation}. Our work instead asks whether the InfoNCE identification principle can be applied along the feature axis.

\textbf{Redundancy Reduction and Feature Structure.}
A complementary family of methods regulates the structure of learned features. Barlow Twins~\citep{zbontar2021barlow}, VICReg~\citep{bardes2021vicreg}, and whitening-based methods~\citep{ermolov2021whitening} reduce redundancy through correlation, covariance, or whitening constraints. More recent work has studied dimensional collapse and decorrelation~\citep{sansone2025collapse}, entropy-based feature diversification~\citep{chakraborty2025entropy}, spectral regularization of feature covariance~\citep{ghanooni2025mitigating}, and theoretical properties of whitening~\citep{han2026cca}. In contrast, our feature-wise objective promotes informative and non-redundant features through relative contrastive competition rather than explicit covariance, entropy, whitening, or spectral penalties. 

\textbf{Feature-wise Contrastive Learning.}
Feature-level objectives have been explored in Zero-CL~\citep{zhang2021zero}, ARB~\citep{zhang2022align}, DimCL~\citep{nguyen2023dimcl}, and SimCSE++~\citep{xu2023simcse++}. In particular, DimCL and SimCSE++ apply InfoNCE along the feature axis, yielding objectives closely related to $\mathcal{L}_{\mathrm{feat}}$. Our contribution is instead the mutual-information analysis: we show that feature-wise identification lower-bounds feature-wise mutual information up to a dependence and non-exchangeability penalty $\Delta$, relate both axis-wise quantities to $I(Z;\hat Z)$, and derive the combined bound. We further show that the feature-wise objective is competitive on its own and extends beyond single-modality representation learning. Relatedly, \citet{garrido2023duality} connect sample- and dimension-contrastive learning through covariance-based criteria, whereas our analysis concerns softmax-based identification.

\section{Conclusion} 
We studied sample-wise and feature-wise contrastive learning through a common mutual information formulation. Our analysis characterizes the feature-wise objective in terms of a mutual-information bound that accounts for dependence and non-exchangeability among feature channels, and connects both axis-wise objectives to the mutual information between the full embedding matrices. This provides a principled basis for combining the two objectives. Experiments across representation learning and cross-modal retrieval show that combining both axes improves performance.

{
    \small
    \bibliographystyle{ieeenat_fullname}
    \bibliography{main}
}

\clearpage
\maketitlesupplementary
\setcounter{page}{1}
\appendix
We provide additional analysis, implementation details, and theoretical proofs in the appendices. The appendices are organized as follows: \begin{itemize} \item \textbf{\hyperref[app:complexity]{Appendix A: Computational and Memory Complexity.}} Analysis of the computational and memory costs of the sample-wise, feature-wise, and combined objectives. \item \textbf{\hyperref[app:implementation]{Appendix B: Implementation Details.}} Complete architectural, optimization, pretraining, and evaluation details for the experiments. \item \textbf{\hyperref[app:proofs]{Appendix C: Proofs.}} Proofs of the sample-wise and feature-wise mutual information bounds, as well as the relation between the axis-wise and joint mutual information, and feature-wise dependence penalty analysis. \item \textbf{\hyperref[app:delta-synthetic]{Appendix D: Synthetic Analysis of the Feature-Wise Dependence Penalty.}} Controlled Gaussian experiments illustrating the behavior of $\Delta$ under feature dependence and non-exchangeability.\end{itemize}

\section{Computational and Memory Complexity}
\label{app:complexity}

We analyze the computational and memory costs of the sample-wise, feature-wise, and combined objectives. We focus on the cost of computing the contrastive objectives themselves, excluding the shared encoder and projection network.

\paragraph{Sample-wise objective.}
The sample-wise objective computes pairwise similarities among the $B$ sample representations across the two branches. Since each similarity is computed between $D$-dimensional vectors, evaluating all $B^2$ pairwise similarities requires $\mathcal{O}(B^2D)$ operations, while explicitly materializing the corresponding logits requires $\mathcal{O}(B^2)$ memory. The row-wise normalization of the representations requires only $\mathcal{O}(BD)$ computation and therefore does not affect the leading-order complexity.

\paragraph{Feature-wise objective.}
The feature-wise objective instead computes pairwise similarities among the $D$ feature channels across the two branches. Since each similarity is computed between vectors of length $B$, evaluating all $D^2$ pairwise similarities requires $\mathcal{O}(BD^2)$ operations, while explicitly materializing the corresponding logits requires $\mathcal{O}(D^2)$ memory. The feature standardization and normalization performed before computing the similarities require only $\mathcal{O}(BD)$ computation and therefore do not affect the leading-order complexity.

\paragraph{Combined objective.}
The combined objective evaluates both the sample-wise and feature-wise losses, resulting in a total similarity-computation cost of
\begin{equation}
\mathcal{O}(B^2D + BD^2)
=
\mathcal{O}\big(BD(B+D)\big).
\end{equation}
If the two logit matrices are materialized simultaneously, their memory cost is $\mathcal{O}(B^2+D^2)$. Alternatively, the two losses can be evaluated sequentially, in which case the two logit matrices need not coexist in memory.

Table~\ref{tab:complexity} summarizes these costs.

\begin{table}[htbp]
\centering
\caption{Computational and memory complexity of the contrastive objectives. The memory cost refers to the similarity logits and excludes the shared encoder and projector activations.}
\label{tab:complexity}
\renewcommand{\arraystretch}{1.15}
\begin{tabular}{lcc}
\toprule
Objective & Similarity computation & Logit memory \\
\midrule
$\mathcal{L}_{\mathrm{sam}}$
& $\mathcal{O}(B^2D)$
& $\mathcal{O}(B^2)$ \\

$\mathcal{L}_{\mathrm{feat}}$
& $\mathcal{O}(BD^2)$
& $\mathcal{O}(D^2)$ \\

$\mathcal{L}_{\mathrm{comb}}$
& $\mathcal{O}(B^2D+BD^2)$
& $\mathcal{O}(B^2+D^2)$ \\
\bottomrule
\end{tabular}
\end{table}

Under our default configuration, $B=256$ and $D=8192$. The sample-wise objective therefore produces $B^2=65{,}536$ logits, whereas the feature-wise objective produces $D^2=67{,}108{,}864$ logits. When stored densely, these require approximately $0.125$ MiB and $128$ MiB, respectively, in FP16/BF16, or $0.25$ MiB and $256$ MiB in FP32.

The relative arithmetic cost of the feature-wise and sample-wise similarity computations is $BD^2/(B^2D)=D/B$, which equals $32$ under our default configuration. This factor applies only to the computation of the contrastive similarities and not to the total training cost, since the encoder and projection network are shared across objectives and account for a substantial fraction of the end-to-end computation. We therefore additionally report measured training time and peak device memory in Table~\ref{tab:imagenet-results}.

\section{Implementation Details}
\label{app:implementation}

\subsection{ImageNet Pretraining}

\paragraph{Architecture.}
We use a ResNet-50 backbone with the final classification layer removed, yielding a 2048-dimensional representation after global average pooling. The backbone is initialized from scratch. The projection head consists of two linear layers, with a 2048-dimensional hidden layer followed by batch normalization and ReLU. For our sample-wise, feature-wise, and combined objectives, the output dimension of the projection head is set to $D=8192$. Batch normalization is also applied to the output of the projection head, without a subsequent nonlinearity.

\paragraph{Data augmentation.}
For each image, we independently generate two augmented views following the standard SimCLR augmentation pipeline~\citep{chen2020simple}. Each view is obtained using a random resized crop to $224\times224$ pixels with crop scale in $[0.08,1.0]$, followed by random horizontal flipping with probability $0.5$. Color jittering is applied with probability $0.8$, using brightness, contrast, saturation, and hue strengths of $0.8$, $0.8$, $0.8$, and $0.2$, respectively. Images are converted to grayscale with probability $0.2$, and Gaussian blur is applied with probability $0.5$ using a standard deviation sampled uniformly from $[0.1,2.0]$. Finally, images are normalized using the ImageNet channel-wise mean $(0.485,0.456,0.406)$ and standard deviation $(0.229,0.224,0.225)$.

\paragraph{Optimization.}
We pretrain all models for 100 epochs with a global batch size of $B=256$. Optimization is performed using LARS with momentum $0.9$ and weight decay $10^{-6}$. Batch-normalization parameters and biases are excluded from weight decay. We use square-root learning-rate scaling, with learning rate
\[
\eta = 0.075\sqrt{B},
\]
which gives $\eta=1.2$ for $B=256$. The learning rate is linearly warmed up during the first 10 epochs and subsequently decayed to zero using a cosine schedule. Training is performed using mixed-precision computation, and synchronized batch normalization is used for distributed training.

\paragraph{Objectives.}
Unless otherwise specified, the temperature is set to $\tau=0.1$ for both the sample-wise and feature-wise objectives. Before computing the sample-wise objective, sample representations are $\ell_2$-normalized. For the feature-wise objective, each feature channel is first standardized across the batch to have zero mean and unit variance and is subsequently $\ell_2$-normalized. For the combined objective, we use $\pi=0.5$, such that the two losses receive equal weight. The default feature dimension is $D=8192$; the experiments studying feature dimensionality vary only $D$ while keeping the remaining training configuration fixed.

\subsection{ImageNet Evaluation}

All evaluation protocols use the representation produced by the ResNet-50 backbone before the projection head.

\paragraph{$k$-NN evaluation.}
For $k$-NN evaluation, the pretrained backbone is kept fixed. Images from both the training and validation sets are resized to 256 pixels, center-cropped to $224\times224$, and normalized using the ImageNet statistics. We construct the feature bank using the ImageNet training set and classify each validation representation using its $k=200$ nearest neighbors. Neighbor votes are weighted according to their cosine similarity using a temperature of $0.1$. We report Top-1 and Top-5 accuracy on the ImageNet validation set.

\paragraph{Linear evaluation.}
For linear evaluation, the pretrained backbone is frozen and only a linear classifier is optimized. During training, images are randomly resized and cropped to $224\times224$ and randomly flipped horizontally. Validation images are resized to 256 pixels and center-cropped to $224\times224$. The classifier is trained for 90 epochs using SGD with momentum $0.9$, zero weight decay, and a learning rate of $0.1$ for a global batch size of 256. The learning rate is scaled linearly with the batch size and decayed using a cosine schedule without warmup. We report the best Top-1 and Top-5 validation accuracy over training.

\paragraph{Full fine-tuning.}
For full fine-tuning, both the pretrained backbone and a newly initialized linear classification head are optimized using ImageNet labels. We use the same training and validation transformations as in linear evaluation. Fine-tuning is performed for 30 epochs using SGD with momentum $0.9$, zero weight decay, and an initial learning rate of $0.05$ for a global batch size of 256. The learning rate is scaled linearly with batch size and follows a cosine schedule without warmup. We report the best Top-1 and Top-5 validation accuracy.





\subsection{Cross-Modal Retrieval}

For Flickr30K, we use a frozen ViT-L/14 image encoder and a frozen BERT-base text encoder. Only linear projection heads are trained, mapping the two modalities to a common $D=8192$-dimensional representation space.

The projection heads are trained for 30 epochs with a batch size of 256 using an optimizer and learning rate similar to those used for the ImageNet benchmark. We use $\tau=0.1$ and $\pi=0.5$ for the contrastive objectives. At test time, image and text embeddings are ranked using cosine similarity. We evaluate using the standard Flickr30K test split and report Recall@1, Recall@5, and Recall@10 for both image-to-text and text-to-image retrieval.

\subsection{Implementation and Hardware}

Our implementation is based on PyTorch and uses distributed data-parallel training. ImageNet experiments use 16-bit mixed precision and synchronized batch normalization across devices. The global batch size is 256. Experiments are conducted using 2 NVIDIA V100 GPUs with 32 GB of memory each.

\section{Proofs} \label{app:proofs}

This section contains the proofs of the theoretical results used in the main text:
\begin{itemize}
    \item Proof of the sample-wise mutual information inequality in Equation~\ref{eq:barber-agakov}.
    \item Proof of the feature-wise mutual information inequality in Equation~\ref{eq:barber-agakov-feature}.
    \item Proof of Lemma~\ref{lem:axis-lower-bound}, relating the axis-wise mutual information quantities to the joint mutual information.
    \item Gaussian characterization of the feature-wise dependence penalty $\Delta$ and sufficient conditions under which it is small.
\end{itemize}

\subsection{Proof of the left inequality in~\eqref{eq:barber-agakov}}
\label{app:proof-outer-ineq}

We show that
\[
I(z_i;\hat z_i)
\;\ge\;
I(U;z_i,X)
\]
holds as a direct consequence of the chain rule for mutual information. Fix an anchor $z_i$, jointly distributed with its true partner $\hat z_i$ as $(z_i,\hat z_i)\sim p(z,\hat z)$, and let $U\in[B]$ be a uniform index, $p(U=u)=1/B$, drawn independently of $(z_i,\hat z_i)$. Write $X=(X_1,\ldots,X_B)$ for the candidate tuple, generated conditionally on $(z_i,U=u)$ by placing the true partner in slot $u$,
\[
X_u\sim p(\hat z\mid z_i),
\]
and drawing the remaining candidates independently from the marginal,
\[
X_n\sim p(\hat z),
\qquad n\neq u.
\]
The $B-1$ negative candidates are drawn independently of $z_i$, of $X_u$, and of each other.

\textbf{Expansion A (condition on $U$ first).}
By the chain rule,
\begin{equation}
I(z_i;X,U)
=
I(z_i;U)
+
I(z_i;X\mid U).
\label{eq:app-expansion-a}
\end{equation}
Since $U$ is drawn independently of $(z_i,\hat z_i)$, $I(z_i;U)=0$. Conditioned on $U=u$, the tuple $X$ consists of the true partner $X_u$ together with $B-1$ negatives that are independent of $z_i$. Therefore,
\begin{align}
I(z_i;X\mid U)
&=
I(z_i;X_U\mid U)
+
I(z_i;X_{-U}\mid X_U,U) \\
&=
I(z_i;\hat z_i),
\end{align}
where the second term vanishes by the independence of the negative candidates. Substituting into Equation~\ref{eq:app-expansion-a} gives
\begin{equation}
I(z_i;X,U)
=
I(z_i;\hat z_i).
\label{eq:app-a-result}
\end{equation}

\textbf{Expansion B (condition on $X$ first).}
By the chain rule again,
\begin{equation}
I(z_i;X,U)
=
I(z_i;X)
+
I(U;z_i\mid X),
\label{eq:app-expansion-b}
\end{equation}
where we use the symmetry of conditional mutual information.

\textbf{Identifying $I(U;z_i,X)$.}
Applying the chain rule to $I(U;z_i,X)$ gives
\begin{equation}
I(U;z_i,X)
=
I(U;X)
+
I(U;z_i\mid X).
\label{eq:app-uzx}
\end{equation}
Marginalizing over $z_i$, the candidate in the true-partner slot also follows the marginal distribution $p(\hat z)$. Therefore,
\[
p(X\mid U=u)
=
\prod_{n=1}^{B}p(X_n),
\]
which does not depend on $u$. Hence, $U$ and $X$ are marginally independent and $I(U;X)=0$. Equation~\ref{eq:app-uzx} therefore gives
\[
I(U;z_i,X)
=
I(U;z_i\mid X),
\]
and Equation~\ref{eq:app-expansion-b} becomes
\begin{equation}
I(z_i;X,U)
=
I(z_i;X)
+
I(U;z_i,X).
\label{eq:app-b-result}
\end{equation}

\textbf{Combining.}
Equating Equations~\ref{eq:app-a-result} and~\ref{eq:app-b-result} gives
\begin{equation}
I(z_i;\hat z_i)
=
I(z_i;X)
+
I(U;z_i,X).
\label{eq:app-exact-identity}
\end{equation}
Since mutual information is non-negative, $I(z_i;X)\ge0$, and therefore
\[
I(z_i;\hat z_i)
\;\ge\;
I(U;z_i,X),
\]
which proves the left inequality in Equation~\ref{eq:barber-agakov}.

\subsection{Proof of the left inequality in~\eqref{eq:barber-agakov-feature}}
\label{app:proof-noniid-penalty}

We show that
\[
I(c_j;\hat c_j)
\;\ge\;
I(V;c_j,Y)-\Delta_j,
\]
where the feature-wise dependence penalty is
\[
\Delta_j
=
I\big(c_j;Y_{-V}\mid Y_V,V\big)
+
I(V;Y),
\]
as a consequence of the chain rule for mutual information and the non-negativity of mutual information.

Fix an anchor feature $c_j$ and its corresponding partner $\hat c_j$. Let $V\in[D]$ denote the position of the true partner after independently and uniformly relabeling the $D$ candidate channels, and let $Y=(Y_1,\ldots,Y_D)$ denote the resulting candidate tuple. By construction, $Y_V=\hat c_j$, $V\sim\mathrm{Unif}[D]$, and $V$ is independent of the underlying feature representations prior to relabeling.

For convenience, define
\begin{equation}
A_j
:=
I\big(c_j;Y_{-V}\mid Y_V,V\big),
\label{eq:app-feature-dependence}
\end{equation}
so that
\begin{equation}
\Delta_j
=
A_j+I(V;Y).
\label{eq:app-feature-penalty}
\end{equation}

\textbf{Expansion A (condition on $V$ first).}
By the chain rule,
\begin{equation}
I(c_j;Y,V)
=
I(c_j;V)
+
I(c_j;Y\mid V).
\label{eq:app-feat-expansion-a}
\end{equation}
Since the relabeling is independent of the feature representations, $I(c_j;V)=0$. Applying the chain rule to the second term gives
\begin{align}
I(c_j;Y\mid V)
&=
I(c_j;Y_V\mid V)
+
I(c_j;Y_{-V}\mid Y_V,V) \\
&=
I(c_j;\hat c_j)+A_j,
\label{eq:app-feat-cond-split}
\end{align}
where the second equality follows because $Y_V$ is the true partner $\hat c_j$ and the random relabeling is independent of $(c_j,\hat c_j)$. Therefore,
\begin{equation}
I(c_j;Y,V)
=
I(c_j;\hat c_j)+A_j.
\label{eq:app-feat-a-result}
\end{equation}

\textbf{Expansion B (condition on $Y$ first).}
By the chain rule,
\begin{equation}
I(c_j;Y,V)
=
I(c_j;Y)
+
I(V;c_j\mid Y),
\label{eq:app-feat-expansion-b}
\end{equation}
where we use the symmetry of conditional mutual information. Applying the chain rule to $I(V;c_j,Y)$ gives
\begin{equation}
I(V;c_j,Y)
=
I(V;Y)
+
I(V;c_j\mid Y).
\label{eq:app-feat-vcjy}
\end{equation}
Rearranging,
\begin{equation}
I(V;c_j\mid Y)
=
I(V;c_j,Y)-I(V;Y).
\label{eq:app-feat-cond-v}
\end{equation}
Substituting Equation~\ref{eq:app-feat-cond-v} into Equation~\ref{eq:app-feat-expansion-b} gives
\begin{equation}
I(c_j;Y,V)
=
I(c_j;Y)
+
I(V;c_j,Y)
-
I(V;Y).
\label{eq:app-feat-b-result}
\end{equation}

\textbf{Combining.}
Equating Equations~\ref{eq:app-feat-a-result} and~\ref{eq:app-feat-b-result} gives
\begin{equation}
I(c_j;\hat c_j)+A_j
=
I(c_j;Y)
+
I(V;c_j,Y)
-
I(V;Y).
\label{eq:app-feat-exact-identity}
\end{equation}
Rearranging,
\begin{equation}
I(c_j;\hat c_j)
=
I(c_j;Y)
+
I(V;c_j,Y)
-
A_j
-
I(V;Y).
\end{equation}
Using Equation~\ref{eq:app-feature-penalty},
\begin{equation}
I(c_j;\hat c_j)
=
I(c_j;Y)
+
I(V;c_j,Y)
-
\Delta_j.
\end{equation}
Since $I(c_j;Y)\ge0$, we obtain
\[
I(c_j;\hat c_j)
\;\ge\;
I(V;c_j,Y)-\Delta_j,
\]
which proves the left inequality in Equation~\ref{eq:barber-agakov-feature}.

The two components of $\Delta_j$ capture the two departures from the conventional sample-wise construction. The term
$I(c_j;Y_{-V}\mid Y_V,V)$ vanishes when the negative channels are conditionally independent of the anchor given its true partner and index, while $I(V;Y)$ vanishes when the candidate tuple contains no information about the location of the true partner. Thus, both terms vanish under the corresponding conditional-independence and exchangeability conditions.

\subsection{Proof of Lemma~\ref{lem:axis-lower-bound}}
\label{app:proof-coverage}

We first prove the sample-wise inequality. For every $i\in[B]$, $z_i=Z_{i,:}$ is a deterministic function of $Z$, while $\hat z_i=\hat Z_{i,:}$ is a deterministic function of $\hat Z$. Therefore, by the data-processing inequality,
\begin{equation}
I(z_i;\hat z_i)
\;\le\;
I(Z;\hat Z),
\qquad
\text{for every } i\in[B].
\label{eq:app-sample-dpi}
\end{equation}
Averaging over the sample indices and using the definition of $I_{\mathrm{sam}}(Z;\hat Z)$ gives
\begin{align}
I_{\mathrm{sam}}(Z;\hat Z)
&=
\frac{1}{B}
\sum_{i=1}^{B}
I(z_i;\hat z_i) \\
&\le
\frac{1}{B}
\sum_{i=1}^{B}
I(Z;\hat Z) \\
&=
I(Z;\hat Z).
\end{align}

The feature-wise inequality follows analogously. For every $j\in[D]$, $c_j=Z_{:,j}$ is a deterministic function of $Z$, while $\hat c_j=\hat Z_{:,j}$ is a deterministic function of $\hat Z$. Therefore,
\begin{equation}
I(c_j;\hat c_j)
\;\le\;
I(Z;\hat Z),
\qquad
\text{for every } j\in[D].
\label{eq:app-feature-dpi}
\end{equation}
Averaging over the feature indices and using the definition of $I_{\mathrm{feat}}(Z;\hat Z)$ gives
\begin{align}
I_{\mathrm{feat}}(Z;\hat Z)
&=
\frac{1}{D}
\sum_{j=1}^{D}
I(c_j;\hat c_j) \\
&\le
\frac{1}{D}
\sum_{j=1}^{D}
I(Z;\hat Z) \\
&=
I(Z;\hat Z).
\end{align}
Therefore,
\[
I(Z;\hat Z)
\ge
I_{\mathrm{sam}}(Z;\hat Z),
\qquad
I(Z;\hat Z)
\ge
I_{\mathrm{feat}}(Z;\hat Z).
\qquad\blacksquare
\]

\subsection{Gaussian Characterization of the Feature-Wise Dependence Penalty}
\label{app:gaussian-delta}

We provide a sufficient-condition analysis for the feature-wise dependence penalty
\[
\Delta_j
=
I(c_j;Y_{-V}\mid Y_V,V)
+
I(V;Y)
\]
under a Gaussian representation model. Suppose that the rows
$(z_i,\hat z_i)$ of $(Z,\hat Z)$ are i.i.d.\ jointly Gaussian, and let
\[
\Sigma_{\hat z}:=\operatorname{Cov}(\hat z_i)
\]
denote the within-branch covariance of a row representation. Since the rows are identically distributed, $\Sigma_{\hat z}$ does not depend on $i$.

\paragraph{Conditional-dependence term.}
By construction, $Y_V=\hat c_j$, while conditional on $V$, $Y_{-V}$ is a re-indexing of the $D-1$ non-corresponding feature channels. Define
\[
\hat C_{-j}:=(\hat c_d)_{d\neq j}.
\]
Since mutual information is invariant to bijective re-indexing,
\begin{equation}
I(c_j;Y_{-V}\mid Y_V,V)
=
I(c_j;\hat C_{-j}\mid\hat c_j).
\label{eq:app-delta-reindex}
\end{equation}

For any sample index $i$, define
\begin{equation}
R_j^2
:=
1-
\frac{
\operatorname{Var}(Z_{i,j}\mid\hat z_i)
}{
\operatorname{Var}(Z_{i,j}\mid\hat Z_{i,j})
}.
\label{eq:app-partial-r2}
\end{equation}
Because the rows are identically distributed, this quantity does not depend on $i$. It measures the reduction in the residual variance of $Z_{i,j}$ obtained by conditioning on the non-corresponding features $\hat Z_{i,-j}$ in addition to its matched feature $\hat Z_{i,j}$.

Since the $B$ rows are independent, conditional mutual information is additive across samples:
\begin{align}
I(c_j;\hat C_{-j}\mid\hat c_j)
&=
\sum_{i=1}^{B}
I(Z_{i,j};\hat Z_{i,-j}\mid\hat Z_{i,j}) \\
&=
B\,
I(Z_{i,j};\hat Z_{i,-j}\mid\hat Z_{i,j}).
\end{align}
For jointly Gaussian variables,
\begin{align}
I(Z_{i,j};\hat Z_{i,-j}\mid\hat Z_{i,j})
&=
\frac{1}{2}
\log
\frac{
\operatorname{Var}(Z_{i,j}\mid\hat Z_{i,j})
}{
\operatorname{Var}(Z_{i,j}\mid\hat z_i)
} \\
&=
-\frac{1}{2}\log(1-R_j^2).
\end{align}
Combining the above expressions gives
\begin{equation}
I(c_j;Y_{-V}\mid Y_V,V)
=
-\frac{B}{2}\log(1-R_j^2).
\label{eq:app-gaussian-dependence}
\end{equation}

\paragraph{Non-exchangeability term.}
Let
\[
P_{\hat z}
=
\mathcal{N}(0,\Sigma_{\hat z})
\]
denote the Gaussian distribution of a row representation in the second branch, and consider the isotropic reference distribution
\[
Q_{\hat z}
=
\mathcal{N}(0,I_D).
\]
The latter is exchangeable across feature coordinates. For the $B$ independent rows,
\begin{equation}
D_{\mathrm{KL}}
\left(
P_{\hat Z}\,\|\,Q_{\hat Z}
\right)
=
B\,
D_{\mathrm{KL}}
\left(
P_{\hat z}\,\|\,Q_{\hat z}
\right).
\label{eq:app-batch-kl}
\end{equation}

The random relabeling used to construct $(V,Y)$ from the feature channels defines a stochastic mapping applied identically under the two distributions. By the data-processing inequality for KL divergence,
\begin{equation}
D_{\mathrm{KL}}
\left(
P_{V,Y}\,\|\,Q_{V,Y}
\right)
\le
D_{\mathrm{KL}}
\left(
P_{\hat Z}\,\|\,Q_{\hat Z}
\right).
\label{eq:app-delta-kl-dpi}
\end{equation}
Under the isotropic reference distribution, the feature coordinates are exchangeable, so the distribution of $Y$ does not depend on the position $V$ of the matched feature. Hence,
\[
Q_{V,Y}=P_VQ_Y.
\]
Using the chain rule for KL divergence,
\begin{align}
D_{\mathrm{KL}}
\left(
P_{V,Y}\,\|\,P_VQ_Y
\right)
&=
I(V;Y)
+
D_{\mathrm{KL}}(P_Y\|Q_Y) \\
&\ge
I(V;Y).
\end{align}
Combining this inequality with Equation~\ref{eq:app-delta-kl-dpi} gives
\begin{equation}
I(V;Y)
\le
D_{\mathrm{KL}}
\left(
P_{\hat Z}\,\|\,Q_{\hat Z}
\right).
\end{equation}

For zero-mean Gaussian distributions,
\[
D_{\mathrm{KL}}
\left(
\mathcal{N}(0,\Sigma_{\hat z})
\,\middle\|\,
\mathcal{N}(0,I_D)
\right)
=
\frac{1}{2}
\left(
\operatorname{tr}(\Sigma_{\hat z})
-
\log\det\Sigma_{\hat z}
-
D
\right).
\]
Therefore,
\begin{equation}
I(V;Y)
\le
\frac{B}{2}
\left(
\operatorname{tr}(\Sigma_{\hat z})
-
\log\det\Sigma_{\hat z}
-
D
\right).
\label{eq:app-gaussian-exchangeability}
\end{equation}

Combining Equations~\ref{eq:app-gaussian-dependence} and
\ref{eq:app-gaussian-exchangeability} yields
\begin{equation}
\Delta_j
\le
-\frac{B}{2}\log(1-R_j^2)
+
\frac{B}{2}
\left(
\operatorname{tr}(\Sigma_{\hat z})
-
\log\det\Sigma_{\hat z}
-
D
\right).
\label{eq:app-gaussian-delta}
\end{equation}

If the feature coordinates are standardized to have unit marginal variance, then
\[
\operatorname{diag}(\Sigma_{\hat z})=\mathbf{1},
\qquad
\operatorname{tr}(\Sigma_{\hat z})=D.
\]
Equation~\ref{eq:app-gaussian-delta} therefore reduces to
\begin{equation}
\Delta_j
\le
-\frac{B}{2}\log(1-R_j^2)
-
\frac{B}{2}\log\det\Sigma_{\hat z}.
\label{eq:app-gaussian-delta-standardized}
\end{equation}
Consequently, $R_j^2=0$ and $\Sigma_{\hat z}=I_D$ are sufficient for
$\Delta_j=0$. The first condition means that the non-corresponding features provide no additional information about $Z_{i,j}$ once its matched feature $\hat Z_{i,j}$ is known, while the second makes the Gaussian feature distribution isotropic and therefore exchangeable across coordinates.

\section{Synthetic Study of the Feature-Wise Dependence Penalty}
\label{app:delta-synthetic}
The feature-wise bound in Equation~\ref{eq:feature-bound} contains the dependence penalty $\Delta$, which is generally not tractable for learned representations. We therefore study synthetic Gaussian representations for which $I(c_j;\hat c_j)$ and the Gaussian characterization of $\Delta_j$ can be computed analytically, while the feature-wise loss can be estimated by Monte Carlo sampling. This allows us to directly compare the certificate $\log D-\mathbb{E}[\mathcal{L}_{\mathrm{feat}}(j)]$ with the corresponding feature-wise mutual information and examine when the penalty $\Delta_j$ is necessary.

\paragraph{Representation model.}
We draw the $B$ rows of $(Z,\hat Z)$ i.i.d.\ from a one-factor Gaussian model. For each sample $i\in[B]$ and feature $d\in[D]$,
\begin{equation}
\hat Z_{i,d}
=
\sqrt{\rho_d}\,g_i
+
\sqrt{1-\rho_d}\,e_{i,d},
\qquad
Z_{i,d}
=
\alpha\,e_{i,d}
+
\gamma\,g_i
+
\sigma\,\varepsilon_{i,d},
\label{eq:app-synthetic-model}
\end{equation}
where $g_i$, $e_{i,d}$, and $\varepsilon_{i,d}$ are independent standard Gaussian variables and $\sigma=1$. The shared factor $g_i$ induces dependence among the features of $\hat Z$, with correlation $\sqrt{\rho_d\rho_{d'}}$ between features $d\neq d'$, while $e_{i,d}$ is specific to feature $d$. Each feature of $\hat Z$ has unit variance, so the representation is standardized by construction. Since the rows are independent,
\begin{equation}
I(c_j;\hat c_j)
=
-\frac{B}{2}\log\big(1-r_j^2\big),
\qquad
r_j
:=
\operatorname{Corr}(Z_{i,j},\hat Z_{i,j}),
\label{eq:app-synthetic-mi}
\end{equation}
and the dependence term in Equation~\ref{eq:app-gaussian-dependence} can similarly be evaluated from the conditional variances in Equation~\ref{eq:app-partial-r2}.

\begin{figure}[htbp]
\centering
\includegraphics[width=\textwidth]{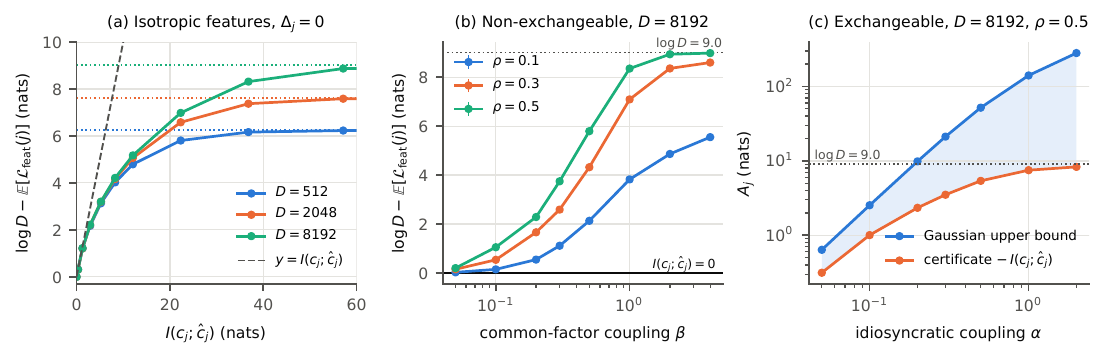}
\caption{Synthetic study of the feature-wise dependence penalty with $B=256$. (a) Isotropic features with $\Delta_j=0$: the certificate $\log D-\mathbb{E}[\mathcal{L}_{\mathrm{feat}}(j)]$ remains below the feature-wise mutual information and saturates at $\log D$ (dotted lines). (b) Non-exchangeable candidates: although $I(c_j;\hat c_j)=0$, the certificate approaches $\log D$. (c) Exchangeable candidates with $\Delta_j=A_j$: the excess of the certificate over the mutual information and the Gaussian characterization bracket $A_j$ as in Equation~\ref{eq:app-synthetic-bracket}.}
\label{fig:synthetic-delta}
\end{figure}

\paragraph{Evaluation.}
For each configuration, we sample minibatches with $B=256$, standardize each feature channel across the minibatch, and compute $\mathcal{L}_{\mathrm{feat}}(j)$ as in Equation~\ref{eq:feature-loss}. When all features are statistically identical, we average over 40 minibatches and 256 anchor features; when a single anchor feature is considered, we average over 400 minibatches. Since the variational bound holds for any scoring function, we report the largest certificate over $\tau\in\{0.05,0.1,0.2,0.5\}$. The Monte Carlo standard errors are below $0.1$ nats in all configurations. Figure~\ref{fig:synthetic-delta} summarizes the three settings considered below.

\paragraph{Isotropic features.}
We first set $\rho_d=0$ and $\gamma=0$, such that the features are independent across channels and $Z_{i,j}$ is a noisy copy of its matched feature $\hat Z_{i,j}$. In this setting, $R_j^2=0$ and $\Sigma_{\hat z}=I_D$, and therefore $\Delta_j=0$ by Equation~\ref{eq:app-gaussian-delta-standardized}. We vary $\alpha$ to control the matched correlation $r_j$ and hence $I(c_j;\hat c_j)$. As shown in Figure~\ref{fig:synthetic-delta}(a), the certificate remains below the feature-wise mutual information across feature dimensions and approaches the $\log D$ ceiling as the mutual information increases. The same behavior is observed for $B\in\{64,256,1024\}$. This verifies the expected behavior of the feature-wise bound when $\Delta_j=0$.

\paragraph{Non-exchangeable candidates.}
We next consider non-exchangeable feature candidates. We set $\rho_j=0$ for the matched feature and $\rho_d=\rho$ for $d\neq j$, such that only the negative candidates depend on the common factor. The anchor is defined as
$Z_{i,j}=-\beta g_i+\sigma\varepsilon_{i,j}$, corresponding to $\alpha=0$ and $\gamma=-\beta$. It is therefore independent of $\hat Z_{i,j}$, giving $I(c_j;\hat c_j)=0$ for every $\beta$, while remaining negatively correlated with the negative candidates through $g_i$. The matched feature can consequently receive the largest relative similarity despite carrying no information about the anchor.

As shown in Figure~\ref{fig:synthetic-delta}(b), the certificate increases with both $\beta$ and $\rho$ and can approach the $\log D$ ceiling. For $D=8192$, it reaches $8.99$ nats compared with $\log D=9.01$; at the default temperature $\tau=0.1$, the corresponding certificate is $6.70$ nats. Thus, the variational term alone can substantially exceed the feature-wise mutual information, and the penalty $\Delta_j$ is required for the bound in Equation~\ref{eq:barber-agakov-feature}. Both components of $\Delta_j$ are present in this construction: the negative candidates contain information about the anchor through the common factor, and the matched candidate is statistically distinguishable from the remaining candidates without observing the anchor. Importantly, this occurs despite all features being standardized, showing that standardization alone does not ensure a small $\Delta_j$.

\paragraph{Exchangeable candidates.}
Finally, we isolate the dependence component of the penalty. We set $\rho_d=\rho$ for every $d$, so that the candidate features are exchangeable and $I(V;Y)=0$. Equation~\ref{eq:feature-penalty} then gives $\Delta_j=A_j$, where $A_j=I(c_j;Y_{-V}\mid Y_V,V)$ is defined in Equation~\ref{eq:app-feature-dependence}. We choose
$\gamma=-\alpha\sqrt{(1-\rho)/\rho}$, which gives
$\operatorname{Cov}(Z_{i,j},\hat Z_{i,j})=0$ and therefore $I(c_j;\hat c_j)=0$ for every $\alpha$. Although the anchor shares the feature-specific component $e_{i,j}$ with its matched feature, this covariance is cancelled by its dependence on the common factor. The matched feature can nevertheless remain identifiable relative to the negative candidates because the softmax in Equation~\ref{eq:feature-loss} is invariant to a common shift of the similarities.

In this setting,
\begin{equation}
\log D-\mathbb{E}\big[\mathcal{L}_{\mathrm{feat}}(j)\big]-I(c_j;\hat c_j)
\;\le\;
A_j
\;\le\;
-\frac{B}{2}\log\big(1-R_j^2\big),
\label{eq:app-synthetic-bracket}
\end{equation}
where the first inequality follows from Equation~\ref{eq:barber-agakov-feature} with $I(V;Y)=0$, and the second follows from Equation~\ref{eq:app-gaussian-dependence}. Figure~\ref{fig:synthetic-delta}(c) shows both quantities for $D=8192$ and $\rho=0.5$. The excess of the certificate over the mutual information increases with $\alpha$ and reaches $8.30$ nats, approaching $\log D$ despite $I(c_j;\hat c_j)=0$. Since $I(V;Y)=0$, this excess arises entirely from the dependence between the anchor and the negative candidates. The Gaussian characterization upper-bounds the excess in all configurations and remains within a factor of $2.0$--$2.5$ for weak coupling ($\alpha\le0.1$).

\paragraph{Summary.}
These experiments illustrate the role of the feature-wise dependence penalty. When $\Delta_j=0$, the certificate remains below the feature-wise mutual information and exhibits the expected $\log D$ ceiling. Subtracting the Gaussian characterization of $\Delta_j$ in Equation~\ref{eq:app-gaussian-delta-standardized} restores a valid lower bound in all configurations. These results support the interpretation of $\Delta$ as an intrinsic component of the feature-wise bound and are consistent with the conditions identified following Equation~\ref{eq:gaussian-delta-standardized}: the penalty is small when non-corresponding features provide little additional information about the anchor and the within-branch covariance is close to isotropic.

\end{document}